# A Hidden Markov Model Based System for Entity Extraction from Social Media English Text at FIRE 2015

Kamal Sarkar
Computer Science & Engineering Dept.
Jadavpur University
Kolkata-700032, India
jukamal2001@yahoo.com

## ABSTRACT
This paper presents the experiments carried out by us at Jadavpur University as part of the participation in FIRE 2015 task: *Entity Extraction from Social Media Text - Indian Languages (ESM-IL)*. The tool that we have developed for the task is based on Trigram Hidden Markov Model that utilizes information like gazetteer list, POS tag and some other word level features to enhance the observation probabilities of the known tokens as well as unknown tokens. We submitted runs for English only. A statistical HMM (Hidden Markov Models) based model has been used to implement our system. The system has been trained and tested on the datasets released for FIRE 2015 task: *Entity Extraction from Social Media Text - Indian Languages (ESM-IL)*. Our system is the best performer for English language and it obtains precision, recall and F-measures of 61.96, 39.46 and 48.21 respectively.

## Categories and Subject Descriptors
H.3 [**Information Storage and Retrieval**]: H.3.1 Content Analysis and Indexing; H.3.3 Information Search and Retrieval; H.3.4 Systems and Software; H.2.3 [Database Management]: Languages-Query Languages

## 1. INTRODUCTION
The objective of named entity recognition is to identify and classify every word/term in a document into some predefined categories like person name, location name, organization name, miscellaneous name (date, time, percentage and monetary expressions etc.) etc.

NER is an important task, having applications in Information Extraction, Question Answering, Machine Translation, Summarization, Cross-lingual information access and other NLP applications. Over the past decade, Indian language content on various social media( twitter, facebook etc.) is rapidly increasing. When the different companies are interested to ascertain public views on their products and services, they need natural language processing software systems which identify entities and relations among the entities. So, there is a need for automatic entity extraction system.

This paper presents a description of HMM (Hidden Markov Model) based system for Entity Extraction from Social Media Text in Indian Languages. This named entity recognition system (NER) considers a variety of entity types: artifact, entertainment, facilities, location, locomotive, materials, organization, person, plants, count, distance, money, quantity, date, day, period, time and year, month, Living thing and Sday.

The task *"Entity Extraction from Social Media Text - Indian Languages (ESM-IL)"* was defined to build the NER systems for four Indian languages - English, Malayalam, Tamil and Hindi for which training data and test data were provided. We have participated for English language only.

The earliest works on named entity recognition (NER) primarily uses two major approaches to NER: Rule based (Linguistic) approaches and Machine Learning (ML) based approaches.

The rule based approaches typically use a set of hand crafted rules [1][2][3].

Machine learning (ML) based techniques for NER make use of a large amount of NE annotated training data to acquire higher level language knowledge from the labeled data. Several ML techniques have already been applied for the NER tasks such as Markov Model (HMM) [4], Maximum Entropy (MaxEnt) [5][6], Conditional Random Field (CRF)[7] etc.

The hybrid approaches that combines different ML approaches are also used. Srihari et al.(2000) [8] combines MaxEnt, Hidden Markov Model (HMM) and handcrafted rules to build an NER system.

NER systems also use gazetteer lists for identifying names. Both the linguistic approach [1][3] and the ML based approach[5][8] may use gazetteer lists.

The NER tasks for Hindi have been presented in [9][10][11].

A discussion on the training data is given in Section 2. The HMM based NER system is described in Section 3. Various features used in NER are then discussed. Next we present the experimental results and related discussions in Section 5. Finally Section 6 concludes the paper.

## 2. TRAINING DATA PREPARATION
The training data released for the FIRE shared task contains two files: one file contains the raw text file and another file contains the NE annotation file in which each row has 6 columns: tweet-id, user-id, NE-tag, NE raw string, NE-start index and NE_length. Index column is the starting character position of NE calculated for each tweet. The participants are instructed to produce the output in the same format after testing the system on the test data. Our system uses the two files supplied for training data and converts the data into the IOB format before training and the data converted in IOB (Inside, Outside and Beginning) format (a format used for the CoNLL-2003 shared task on NER) is used for training. IOB format uses a B−XXX tag that indicates the first word of an entity type XXX and I−XXX that is used for subsequent words of an entity. The tag "O" indicates the word is outside of an NE (i.e., not a part of a named entity).

## 3. HMM BASED NAMED ENTITY TAGGING

A named entity recognizer based on Hidden Markov Model (HMM) finds the best sequence of NE tags $t_1^n$ that is optimal for a given observation sequence $o_1^n$. The tagging problem becomes equivalent to searching for $\arg\max_{t_1^n} P(o_1^n | t_1^n) P(t_1^n)$ (by the application of Bayes' law), that is, we need to compute:

$$\hat{t}_1^n = \arg\max_{t_1^n} P(o_1^n | t_1^n) P(t_1^n) \quad (1).$$

Where $t_1^n$ is a tag sequence and $o_1^n$ is an observation sequence, $P(t_1^n)$ is the prior probability of the tag sequence and $P(o_1^n | t_1^n)$ is the likelihood of the word sequence.

In general, HMM based sequence labeling tasks such as POS tagging use words in a sentence as an observation sequence [12] [13]. But, we use MontyTagger [14] to assign POS tags to the data released for the task, that is, some additional information such as POS for each token in a tweet becomes now available. We also use some other information such as whether the token contains any digit, whether the token contains any hash tag or not etc. We use this information in a form of meta tag (details are presented in the subsequent sections). We use gazetteer information also. If any token is found in the specific gazetteer list, we use the gazetteer tag in place of POS tag (details are presented in the subsequent sections).

Unlike the traditional HMM based NER system, to use this additional information for named entity recognition task, we consider a triplet as an observation symbol: <word, POS-tag/gazetteer tag, meta-tag>. This is a pseudo token used as an observed symbol, that is, for a tweet of *n* words, the corresponding observation sequence will be as follows:

(<word_1, X-tag_1, meta-tag_1>, <word_2, X-tag_2, meta-tag_2>, <word_3, X-tag_3, meta-tag_3>, .........., <word_n, X-tag_n, meta-tag_n>). Here an observation symbol $o_i$ corresponds to <word_i, X-tag_i, meta-tag_i> and X-tag can be either POS tag or gazetteer tag).

Since Equation (1) is too hard to compute directly, HMM taggers follows Markov assumption according to which the probability of a tag is dependent only on short memory (a small, fixed number of previous tags). For example, a bigram tagger considers that the probability of a tag depends only on the previous tag

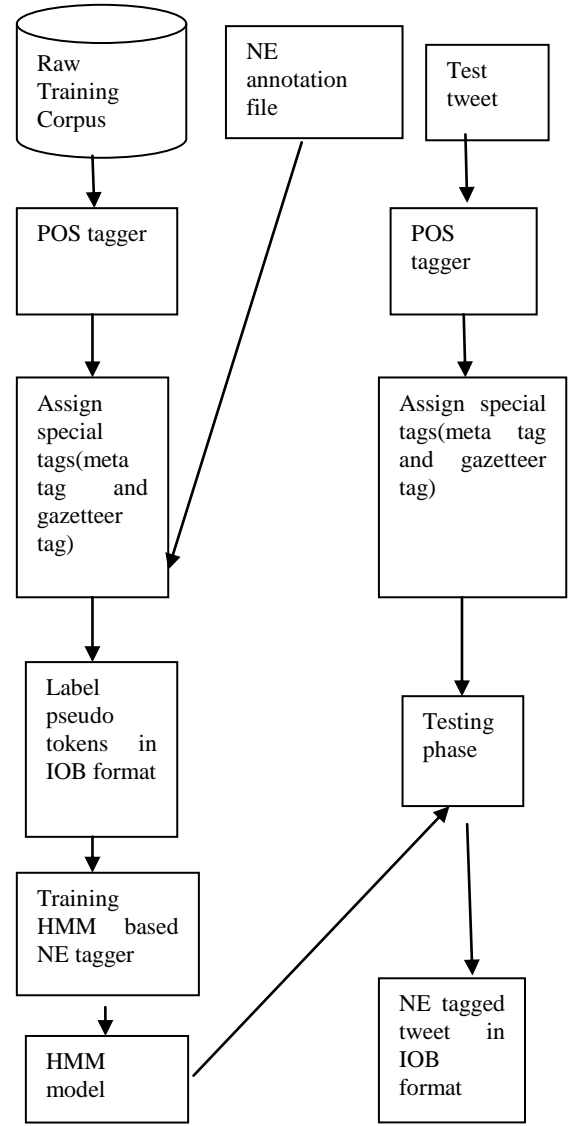

**Figure 1. Architecture for our developed HMM based NE extraction system**

For our proposed trigram model, the probability of a tag depends on two previous tags and thus $P(t_1^n)$ is computed as:

$$P(t_1^n) \approx \prod_{i=1}^{n} P(t_i | t_{i-1}, t_{i-2}) \quad (2)$$

Depending on the assumption that the probability of a word appearing is dependent only on its own tag, $P(o_1^n | t_1^n)$ can be simplified to:

$$P(o_1^n | t_1^n) \approx \prod_{i=1}^{n} P(o_i | t_i) \quad (3)$$

Plugging the above mentioned two equations (2) and (3) into (1) results in the following equation by which a bigram tagger estimates the most probable tag sequence:

$$\hat{t}_1^n = \arg\max_{t_1^n} P(t_1^n | o_1^n) P(t_1^n) \approx \arg\max_{t_1^n} \prod_{i=1}^{n} P(o_i | t_i) P(t_i | t_{i-1}) \quad (4)$$

Where: the tag transition probabilities, $P(t_i | t_{i-1})$, represent the probability of a tag given the previous tag. $P(o_i | t_i)$ represents the probability of an observed symbol given a tag.

Considering a special tag $t_{n+1}$ to indicate the end sentence boundary and two special tags $t_{-1}$ and $t_0$ at the starting boundary of the sentence and adding these three special tags to the tag set [15], gives the following equation for NE tagging:

$$\hat{t}_1^n = \arg\max_{t_1^n} P(t_1^n | o_1^n) P(t_1^n) \approx$$

$$\arg\max_{t_1^n} [\prod_{i=1}^{n} P(o_i | t_i) P(t_i | t_{i-1}, t_{i-2})] P(t_{n+1} | t_n) \quad (5)$$

The equation (5) is still computationally expensive because we need to consider all possible tag sequence of length *n*. So, dynamic programming approach is used to compute the equation (5).

At the training phase of HMM based NE tagging, observation probability matrix and tag transition probability matrix are created. Architecture of our developed NE tagger is shown in Figure 1.

### 3.1 Computing Tag Transition Probabilities

As we can see from the equation (4), to find the most likely tag sequence for an observation sequence, we need to compute two kinds of probabilities: tag transition probabilities and word likelihoods or observation probabilities.

Our developed trigram HMM tagger requires to compute tag trigram probability, $P(t_i | t_{i-1}, t_{i-2})$, which is computed by the maximum likelihood estimate from tag trigram counts. To overcome the data sparseness problem, tag trigram probability is smoothed using deleted interpolation technique [13][15] which uses the maximum likelihood estimates from counts for tag trigram, tag bigram and tag unigram.

### 3.2 Computing Observation Probabilities

The observation probability of a observed triplet <word, X-tag, meta-tag >, which is the observed symbol in our case, is computed using the following equation [12][13].

$$P(o | t) = \frac{C(o,t)}{C(o)} \quad (7)$$

### 3.3 Viterbi Decoding

The task of a decoder is to find the best hidden state sequence given an input HMM and a sequence of observations.

The Viterbi algorithm is the most common decoding algorithm used for HMM based tagging task. This is a standard application of the classic dynamic programming algorithm[16].

Given a tag transition probability matrix and the observation probability matrix, Viterbi decoding (used at the testing phase) accepts a tweet in Indian language and finds the most likely tag sequence for the test tweet which is also X-tagged and Meta tagged. Here a tweet is submitted to the viterbi as the observation sequence of triplets:

(<word$_1$, X-tag$_1$, meta-tag$_1$>, <word$_2$, X-tag$_2$, meta-tag$_2$>, <word$_3$, X-tag$_3$, meta-tag$_3$>, .........., <word$_n$, X-tag$_n$, meta-tag$_n$>) . Here an observation symbol $o_i$ corresponds to <word$_i$, X-tag$_i$, meta-tag$_i$> and X-tag can be either POS tag or gazetteer tag).

After assigning the tag sequence to the observation sequence as mentioned above, X-tag and meta-tag information are removed from the output and thus the output for an input sentence is converted to a NE-tagged sentence.

We have used the Viterbi algorithm presented in [16] for finding the most likely tag sequence for a given observation sequence.

One of the important problems to apply Viterbi decoding algorithm is how to handle unknown triplets in the input. The unknown triplets are triplets which are not present in the training set and hence their observation probabilities are not known. To handle this problem, we estimate the observation probability of an unknown one by analyzing X-tag, meta-tag and the suffix of the word associated with the corresponding the triplet. We estimate the observation probability of an unknown observed triplet in the following ways:

The observation probabilities of unknown triplet < word, X-tag, meta-tag> corresponding to a word in the input sentence are decided according to the suffix of a pseudo word formed by adding X-tag and meta-tag to the end of the word. We find the observation probabilities of such unknown pseudo words using suffix analysis of all rare pseudo words (frequency <=2) in the training corpus for the concerned language [13][15].

## 4. SPECIAL TAGS

### 4.1 Meta Tag

Each token has some properties by which one token differs from another. For example, a token may only consist of digits or it may contain hash. To capture such information specific to a token, we use Meta tag. For example, if a token is consisting of only digits, meta tag that we will assign to the token is ALLDIGITS which we write ALDT in short.

The various meta tags that we use for our task are described below. Meta tag for a token is determined using the following rules which are fired in the following order.

Meta-tag="YYYY"(default)
*if the first letter of the token is a capital letter then
metatag = "ICAP"
end if*

*if the first token is abbreviation then
metatag = "ABBR"
End If*

*if contains "#" at the begining of the token and the first character after hash is a capital letter then
    metatag = "CHAS"
ElseIf contains "#" at the begining of the token Then
metatag = "HASH"
End If*

*if contains "@" at the begining of the token then
  metatag = "ATSY"*

*End If*

*If last charater is a colon(":") And the first letter is capital then*
   *metatag = "CCOL"*
*ElseIf last charater is a colon(":") Then*
   *metatag = "COLN"*
*End If*

*if contains hyphen and the first character is capital then*
   *metatag = "CHYP"*
*ElseIf hyphen occurs after 3 characters from the begining then*
   *metatag = "HYPH"*
*End If*

*if the token is 4 digits then*
   *metatag = "DFOR"*
*ElseIf the token is two digits then*
   *metatag = "DTWO"*
*ElseIf the token is one digit then*
   *metatag = "DONE"*
*ElseIf token contains at least one digit then*
   *metatag = "DIGT"*
*End If*

*If contains one comma and contains at least one digit then*
   *metatag = "DCOM"*
*ElseIf the last character is a comma and first character is capital then*
   *metatag = "CLCO"*
*ElseIf contains one comma at the end of the token then*
   *metatag = "LCOM"*
*ElseIf contains more than one comma and first character is capital then*
   *metatag = "CMCO"*
*End If*

*If token contains all dots then*
   *metatag = "ALDT"*
*End If*

## 4.2 Gazetteer tag

In earlier sections, we have mentioned that POS tag for a token is replaced by a gazetteer tag if the token is found in a particular gazetteer list. if the length of a raw word is greater than equal to 2 , before searching in the gazetteer list, we remove from the token the symbols such as ",",".",":","#" and "@". The description of gazetteer list is shown in Table 1.

**Table 1: Description of Gazetteer lists**

| Gazetteer name | description | Number of entries in the list |
|---|---|---|
| Bperson | List of first names separated from a list of person names | 657 |
| Iperson | List of words representing second names, third names, last names extracted from a list of person names | 563 |
| Blocation | A list of first words extracted from list of location names | 1243 |
| Ilocation | A list of words extracted from a list of location names where a extracted word is not the first word of the location name | 257 |
| facilities | A list of facility names such as school, college etc. | 14 |
| months | A list of English month names | 12 |
| days | A list of English day names | 7 |
| period | A list of words indicating "period" such as "month", "year" etc. | 34 |
| Count expressions | A list of words indicating "count" | 58 |
| Monetary expressions | A list of words indicating monetary expressions such as lakh, crore etc. | 18 |

We follow the following rules for assigning this type of tag to the token:

X-tag=POS-tag (default tag)

*if Token is found in the BPerson list then*
  *X-tag="BPER"*
*elseif Token is found in the IPerson list then*
   *X-tag = "IPER"*
*elseif Token is found in Blocation list then*
   *X-tag="BLOC"*
*elseif Token is found in ILocation list then*
   *X-tag = "ILOC"*
*elseif Token is found in the list of facilities then*
   *X-tag="FACI"*
*elseif Token is found in the list of month names then*
   *X-tag = "MONT"*
*elseif Token is found in the list of day names then*
   *X-tag= "DAYS"*
*elseif Token is found in the list of period indicating expressions then*
   *X-tag = "PERD"*
*elseif Token is found in the list of expression denoting countthen*
   *X-tag = "COUN"*
*elseif Token is found in the list of monetary expressions then*
   *X-tag = "MONY"*
*End If*

## 5. EVALUATION AND RESULTS

We train separately our developed named entity recognizer based on the training data and tune the parameters of our system on the training data for the English language. After learning the tuning parameters, we test our system on the test data for the concerned language. The description of the data for English language is shown in the Table2

After getting the NE-tagged output in IOB format from the HMM model, we observed that the NE tagged output contains some occurrences of a sequence of I-XXXs where the left boundary of each such sequence is a transition from the tag "O" to I-XXXs (but, according to the IOB format, the left boundary of a named entity is a transition from any tag to B-XXX).

**Table2. The description of the data for English language**

| Language | Total of tweets | | NE Types |
|---|---|---|---|
| | Training data | Test data | |
| English | 11003 | 9641 | 21 |

We have also observed that the word sequence to which this type of tag sequence is assigned is not really a named entity. So, considering this as the errors of the model, we replace such a sequence of I-XXXs in the output by a sequence of "o". After applying this post-processing on the output produced by the HMM model, the final output file is generated.

Our developed NER system has been evaluated using the traditional precision (P), recall (R) and F-measure (F). For training, tuning and testing our system, we have used the dataset for English language, released by the organizers of the *ESM-IL task-* FIRE 2015. The organizers of the *ESM-IL task-* FIRE 2015 released the data in two phases: in the first phase, training data is released along with the corresponding NE annotation file. In the second phase, the test data is released and no NE annotation file is provided. The contestants are instructed to generate NE annotation file for test data using their developed systems. NE annotation file for test data was finally sent to the organizers for evaluation. The organizers evaluate the different runs submitted by the various teams and send the official results to the participating teams.

We have shown in Table 3 the results obtained by our submitted run indicated by team id "KSarkar – JU". As we can see from the table, our system outperforms the other systems participated in the ESM-IL task. Table 3 only shows the FIRE 2015 official results for English language only. The overall FIRE 2015 official results for ESM-IL task including all languages are shown in Table 4.

**Table 3. Official results obtained by the various systems participated in the ESM-IL task- FIRE 2015 for English language**

| Teams | | P | R | F |
|---|---|---|---|---|
| Shriya - Amritha | Run1 | 0.08 | 0.064 | 0.071 |
| Sanjay - Amritha | Run1 | 0.057 | 0.028 | 0.038 |
| | Run2 | 0.043 | 0.021 | 0.028 |
| Chintak - LDRP | Run1 | 7.30 | 4.17 | 5.31 |
| | Run2 | 5.35 | 5.67 | 5.50 |
| **KSarkar - JU** | **Run1** | **61.96** | **39.46** | **48.21** |
| Vira - Charotar Univ | Run1 | 4.13 | 3.39 | 3.72 |
| Pallavi - HITS | Run1 | 50.48 | 32.03 | 39.19 |
| | Run2 | 50.21 | 37.06 | 42.64 |
| | Run3 | - | - | - |
| Sombuddha - JU | Run1 | 46.92 | 32.41 | 38.34 |
| | Run2 | 58.09 | 31.85 | 41.15 |
| | Run3 | 49.10 | 31.59 | 38.45 |
| | Run4 | 46.50 | 30.20 | 36.61 |
| | Run5 | 58.09 | 31.85 | 41.15 |

## 6. CONCLUSION

This paper describes a named entity recognition system for Entity Extraction from Social Media Text in English language. The features such as Gazetteer list, POS tag and some other word level features have been introduced into the HMM model. The experimental results show that our system is the best performer among the systems participated in the ESM-IL task for English language. The named entity recognition system has been developed using Visual Basic platform so that a suitable user interface can be designed for the novice users. The system has been designed in such a way that only changing the training corpus in a file can make the system portable to a new Indian language.

Table 4. Language wise official results obtained by the various systems participated in the *ESM-IL task*- FIRE 2015

| Language | | Hindi | | | Tamil | | | Malayalam | | | English | | |
|---|---|---|---|---|---|---|---|---|---|---|---|---|---|
| eams | | P | R | F | P | R | F | P | R | F | P | R | F |
| Shriya - Amritha | Run1 | 2.61 | 1.44 | 1.86 | 0.19 | 0.066 | 0.09 | 0.23 | 0.29 | 0.26 | 0.08 | 0.064 | 0.071 |
| Sanjay - Amritha | Run1 | 2.27 | 0.16 | 0.29 | 0.30 | 0.07 | 0.12 | 0.075 | 0.036 | 0.04 | 0.057 | 0.028 | 0.038 |
| | Run2 | - | - | - | 0.250 | 0.077 | 0.11 | - | - | - | 0.043 | 0.021 | 0.028 |
| Chintak - LDRP | Run1 | 67.11 | 0.76 | 1.51 | - | - | - | - | - | - | 7.30 | 4.17 | 5.31 |
| | Run2 | 74.73 | 46.84 | 57.59 | - | - | - | - | - | - | 5.35 | 5.67 | 5.50 |
| **KSarkar - JU** | **Run1** | - | - | - | - | - | - | - | - | - | **61.96** | **39.46** | **48.21** |
| Vira - Charotar Univ | Run1 | 25.65 | 16.14 | 19.82 | - | - | - | - | - | - | 4.13 | 3.39 | 3.72 |
| Pallavi - HITS | Run1 | 81.21 | 44.57 | 57.55 | 70.42 | 14.13 | 23.54 | - | - | - | 50.48 | 32.03 | 39.19 |
| | Run2 | 80.86 | 44.25 | 57.20 | 64.52 | 22.14 | 32.97 | - | - | - | 50.21 | 37.06 | 42.64 |
| | Run3 | 81.49 | 41.58 | 55.06 | - | - | - | - | - | - | - | - | - |
| Sombuddha - JU | Run1 | err | - | - | - | - | - | - | - | - | 46.92 | 32.41 | 38.34 |
| | Run2 | err | - | - | - | - | - | - | - | - | 58.09 | 31.85 | 41.15 |
| | Run3 | err | - | - | - | - | - | - | - | - | 49.10 | 31.59 | 38.45 |
| | Run4 | - | - | - | - | - | - | - | - | - | 46.50 | 30.20 | 36.61 |
| | Run5 | - | - | - | - | - | - | - | - | - | 58.09 | 31.85 | 41.15 |